\pdfoutput=1

\documentclass[11pt]{article}
\usepackage[preprint]{acl}
\usepackage{times}
\usepackage{latexsym}
\usepackage[T1]{fontenc}
\usepackage[utf8]{inputenc}
\usepackage{microtype}
\usepackage{inconsolata}
\usepackage{graphicx}
\usepackage{float}
\usepackage{booktabs}
\usepackage{multirow}
\usepackage{multicol}
\usepackage{colortbl}
\usepackage{amsmath}
\usepackage[baseline]{euflag}
\usepackage{xspace}
\newcommand{\prodvec}[0]{\textsc{product}\xspace}
\newcommand{\prodcat}[0]{\textsc{product-category}\xspace}
\newcommand{\hazvec}[0]{\textsc{hazard}\xspace}
\newcommand{\hazcat}[0]{\textsc{hazard-category}\xspace}
\newcommand{\yearfeat}[0]{\textsc{year}\xspace}
\newcommand{\monthfeat}[0]{\textsc{month}\xspace}
\newcommand{\dayfeat}[0]{\textsc{day}\xspace}
\newcommand{\country}[0]{\textsc{country}\xspace}

\newcommand{\titlefeat}[0]{\textsc{title}\xspace}
\newcommand{\textfeat}[0]{\textsc{text}\xspace}

\title{SemEval-2025 Task 9: The Food Hazard Detection Challenge}

\author{%
Korbinian Randl,$^{1}$
John Pavlopoulos,$^{1,2,3}$
Aron Henriksson,$^{1}$
Tony Lindgren,$^{1}$
Juli Bakagianni$^{4}$
\\
$^{1}$ Stockholm University, 
Borgarfjordsgatan 12, 164 07 Kista, Sweden\\
\texttt{\{korbinian.randl,ioannis,aronhen,tony\}@dsv.su.se}\\
$^{2}$Athens University of Economics and Business, Greece\\
$^{3}$Archimedes, Athena Research Center, Greece\\
$^{4}$Agroknow, Greece
}

\begin{document}
\maketitle
\begin{abstract}
    In this challenge, we explored text-based food hazard prediction with long tail distributed classes. The task was divided into two subtasks:
    \textbf{(1)} predicting whether a web text implies one of ten food-hazard categories and identifying the associated food category, and 
    \textbf{(2)} providing a more fine-grained classification by assigning a specific label to both the hazard and the product.
    Our findings highlight that large language model-generated synthetic data can be highly effective for oversampling long-tail distributions.
    Furthermore, we find that fine-tuned encoder-only, encoder-decoder, and decoder-only systems achieve comparable maximum performance across both subtasks.
    During this challenge, we gradually released (under \href{https://creativecommons.org/licenses/by-nc-sa/4.0/deed.en}{CC BY-NC-SA 4.0}) a novel set of 6,644 manually labeled food-incident reports. 
\end{abstract}

\section{Introduction}
The Food Hazard Detection Challenge at SemEval 2025 evaluated explainable classification systems for titles of food-incident reports collected from the world wide web. Algorithms like these could, for example, be used to help automated crawlers find and extract food issues from publicly available sources like social media. Since such systems could have a high economic impact (specific food items may need to be recalled, leading to financial damage for the producers), transparency is extremely important. Human experts using data from these crawlers need to be well-informed about how the respective food issues are extracted.

Prior work has shown that a major challenge in food-hazard and food-product classification from text is the large number of possible classes, combined with a long-tail distribution \citep{randl2024cicle}. To address this, we define two subtasks:

\begin{itemize}
    \item \textbf{Subtask 1~(ST1)} focuses on training models for coarse-grained ``category'' prediction.
    \item \textbf{Subtask 2~(ST2)} is a more fine-grained ``vector'' prediction task.
\end{itemize}

A prior SemEval challenge by \citet{kirk-etal-2023-semeval} framed a similar setup as an initial step toward explainability. While this interpretation may be somewhat broad, we recognize that a ``vector'' prediction task is particularly valuable for automated information extraction, as it provides more specific information.

\begin{figure}[t]
    \centering
    \includegraphics[width=\linewidth, trim={0 2.5cm 0 0},clip]{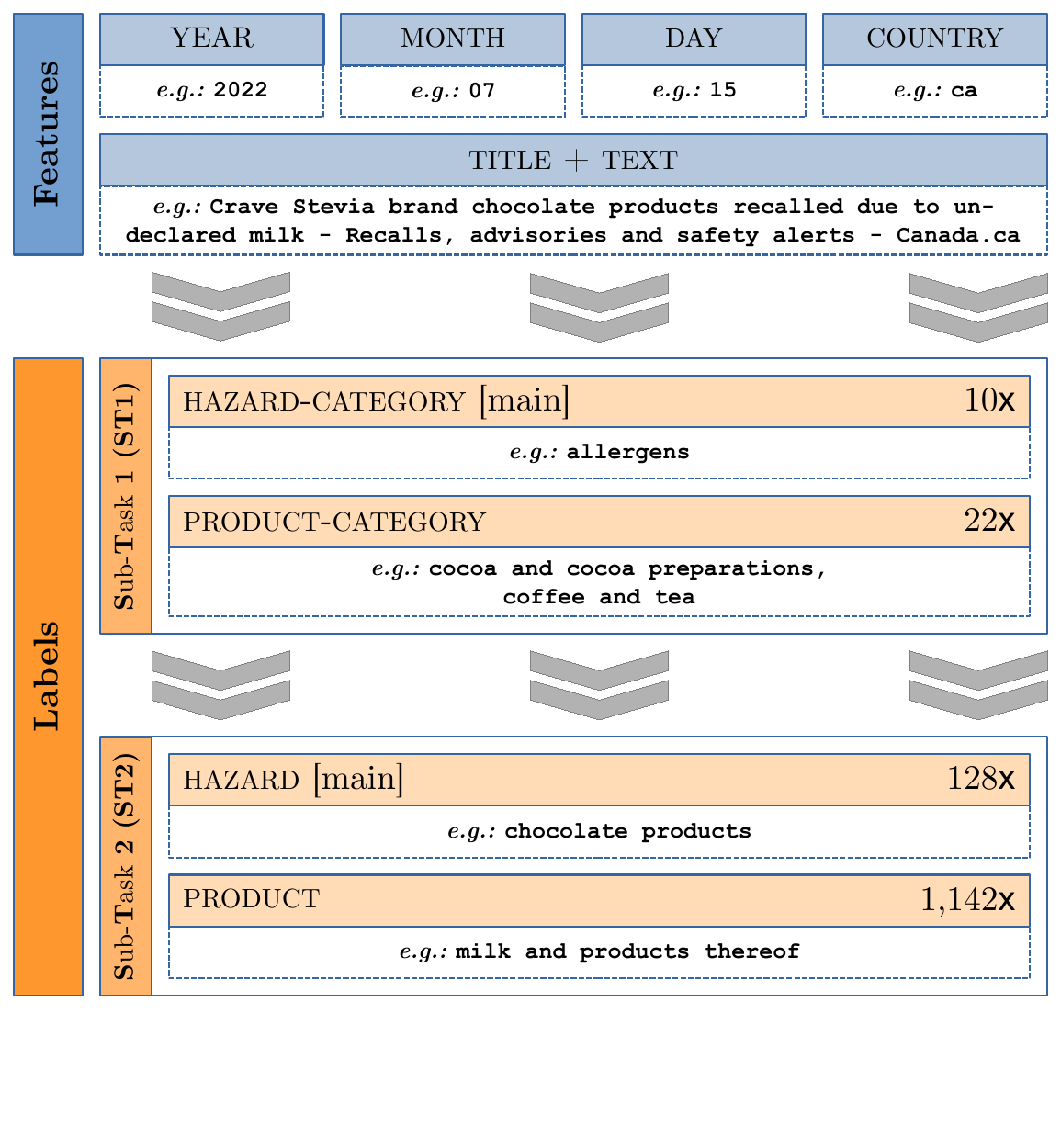}
    \caption{The columns in the blue boxes were available to the participants to serve as model input, while the orange boxes comprised the ground truth labels per sub-task. The number on the right of each label indicated the number of unique values per label.}
    \label{fig:example}
\end{figure}

\begin{figure*}
    \centering
    \includegraphics[width=.8\linewidth]{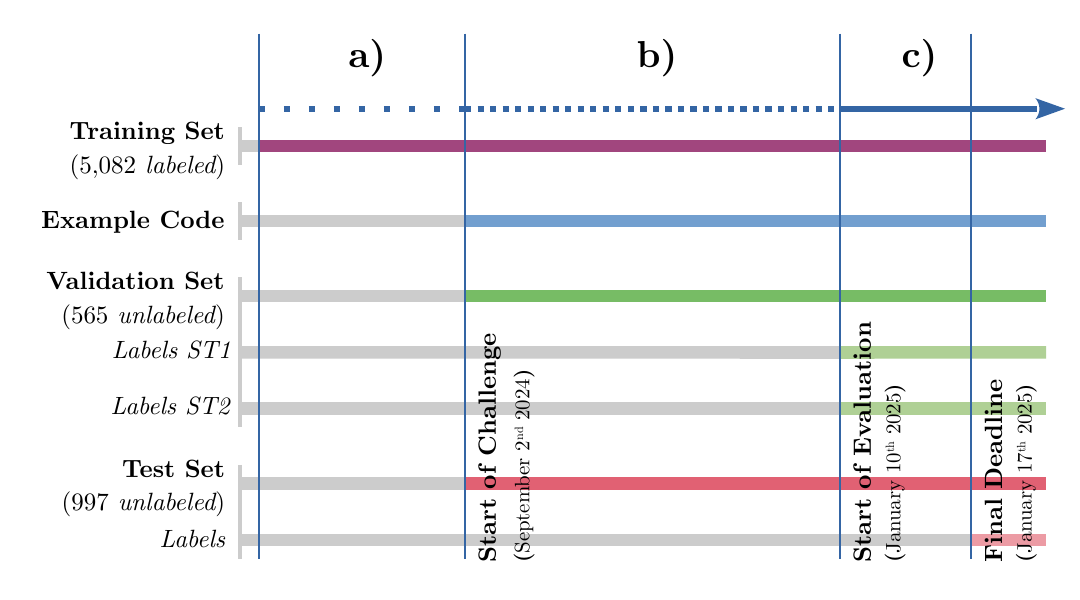}
    \caption{Timeline of the challenge:
        \textbf{(a)~Trial Phase}: Training data was provided before the challenge commenced.
        \textbf{(b)~Conception Phase}: Example code, along with unlabeled validation and test data, was released at the beginning of the challenge. During this phase, participants could submit separate trial entries for ST1 (category classification) and ST2 (``vector'' classification) using the validation data.
        \textbf{(c)~Evaluation Phase}: The validation data was made available, and final submissions for both tasks were accepted on the test data to determine the final ranking.}
    \label{fig:timeline}
\end{figure*}

An overview of the SemEval-Task is shown in Figure~\ref{fig:example}. It includes \textbf{two sub-tasks}: 
\textbf{(ST1)}~text classification for food hazard prediction, predicting the type of hazard (\hazcat) and the type of product (\prodcat);
\textbf{(ST2)}~food hazard and product ``vector'' detection, predicting the exact hazard (\hazvec) and product (\prodvec).
The task was primarily concerned with detecting the hazard (more important than the product), hence a two-step scoring metric based on the macro~$\text{\sc f}_1$ score was used, focusing on the respective hazard label per sub-task (see Section~\ref{sec:eval}).

\section{Task Organization}
The detailed timeline of the project is illustrated in Figure~\ref{fig:timeline}. Participants were provided with training and validation data to develop, train, and evaluate their systems before the evaluation phase. The challenge was conducted on Codalab\footnote{\href{https://codalab.lisn.upsaclay.fr}{\texttt{https://codalab.lisn.upsaclay.fr}}}~\cite{codalab}, adhering to the framework of previous competitions~\citep{kirk-etal-2023-semeval}.

The validation data was made available at the start of the challenge, enabling participants to submit to the leaderboards and compare their systems during the conception phase. However, these rankings did not influence the final results. The test set was released at the beginning of the challenge with labels concealed until its conclusion. During the evaluation phase, models could be trained on both the training and validation data but were evaluated exclusively on the test set to get the final ranking. After the evaluation phase, participants were required to submit a brief system description specifying the dataset features used, with this information made public alongside the final ranking.

Participants could submit up to five times per day and 100 times in total during the conception phase, whereas in the evaluation phase, each participant was limited to a single valid submission. Additionally, participants were required to share their code (e.g., via GitHub) along with their system description papers.

\section{Dataset} \label{sec:data}
\begin{table}[t]
    \centering
    \resizebox*{.45\textwidth}{!}{
    \begin{tabular}{|rl|}
    \hline
    
    \multicolumn{2}{|p{\linewidth}|}{\cellcolor{gray!20}\footnotesize ``Randsland brand Super Salad Kit recalled due to Listeria monocytogenes''} \\
    \hline
    \footnotesize\texttt{\color{orange}\hazvec}:           & \footnotesize listeria monocytogenes \\
    \footnotesize\texttt{\color{orange}\hazcat}:  & \footnotesize biological \\
    \footnotesize\texttt{\color{blue}\prodvec}:            & \footnotesize salads \\
    \footnotesize\texttt{\color{blue}\prodcat}:   & \footnotesize fruits and vegetables \\

    \hline
    \hline
    
    \multicolumn{2}{|p{\linewidth}|}{\cellcolor{gray!20}\footnotesize ``Create Common Good Recalls Jambalaya Products Due To Misbranding and Undeclared Allergens''} \\
    \hline
    \footnotesize\texttt{\color{orange}\hazvec}:           & \footnotesize milk and products thereof \\
    \footnotesize\texttt{\color{orange}\hazcat}:  & \footnotesize allergens \\
    \footnotesize\texttt{\color{blue}\prodvec}:            & \footnotesize meat preparations \\
    \footnotesize\texttt{\color{blue}\prodcat}:   & \footnotesize meat, egg and dairy products \\

    \hline
    \hline
    
    \multicolumn{2}{|p{\linewidth}|}{\cellcolor{gray!20}\footnotesize ``Nestlé Prepared Foods Recalls Lean Cuisine Baked Chicken Meal Products Due to Possible Foreign Matter Contamination''} \\
    \hline
    \footnotesize\texttt{\color{orange}\hazvec}:           & \footnotesize plastic fragment \\
    \footnotesize\texttt{\color{orange}\hazcat}:  & \footnotesize foreign bodies \\
    \footnotesize\texttt{\color{blue}\prodvec}:            & \footnotesize cooked chicken \\
    \footnotesize\texttt{\color{blue}\prodcat}:   & \footnotesize prepared dishes and snacks \\

    \hline
    \end{tabular}
    }
    \caption{Sample of texts along with their labels.}
    \label{tab:example_texts}
\end{table}

\begin{figure*}
    \centering
    \includegraphics[width=\linewidth]{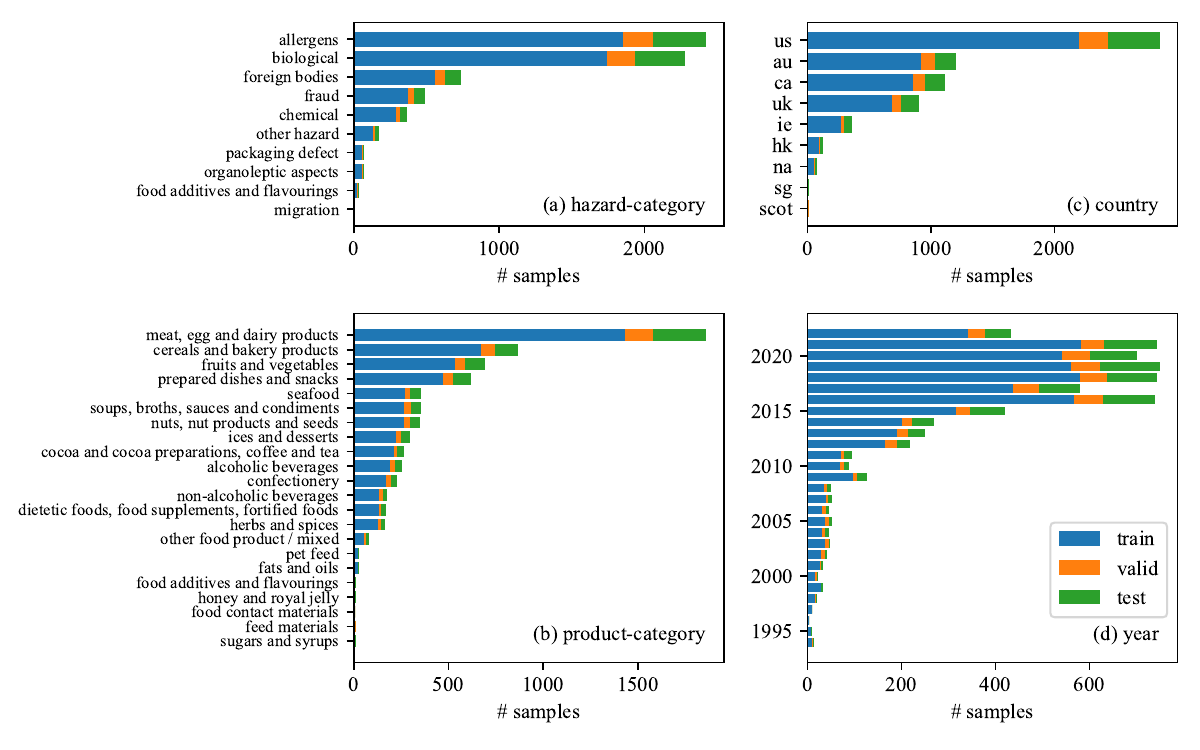}
    \caption{Overview over the data used in the challenge}
    \label{fig:data-stats}
\end{figure*}

\noindent The dataset we used in the challenge is a subset of the data described in \citet{randl2024cicle} and publicly accessible on zenodo~\citep{food_incidents}.
It consists of 6,644 \titlefeat{}s (length in characters: \textit{min=5, avg=88, max=277}), and full \textfeat{}s (length in characters: \textit{min=56, avg=2329, max=48318}) of English food recall announcements from the official websites of food agencies (e.g. the FDA's website). In addition, the dataset contains meta information such as date of download and country of issue.
These texts were primarily gathered between 2012 and 2022 from domains based in the United States, Australia, Canada, and the United Kingdom (see Figure~\ref{fig:data-stats}~(c) and (d)).
The data was manually labeled with the reason for recall (\hazvec) and the recalled \prodvec. Each text has been assessed by two experts on food science or food technology from Agroknow\footnote{\href{https://agroknow.com}{\texttt{https://agroknow.com}}}.
Some sample \titlefeat{}s are shown in Table~\ref{tab:example_texts}.

The data was stratified based on the more important hazard ``vectors'' (\hazvec) and divided into three subsets: 5,082 samples for training, 565 for validation, and 997 for evaluation. 
The training data, which was already published on zenodo~\citep{food_incidents}, also contains additional non-English texts that could be used by participants to train their classifiers. Nevertheless, our evaluation was only based on English texts.
As the texts contain varying degrees of information on the \hazvec, we considered careful pre-processing of the data as part of the challenge.
Upon completion of the task, the complete dataset was made available under the \href{https://creativecommons.org/licenses/by-nc-sa/4.0/deed.en}{Creative Commons BY-NC-SA~4.0} license.

One sample of the dataset is shown in Figure~\ref{fig:example}. As described above, the data includes the features 
\yearfeat, \monthfeat, \dayfeat,
\country, \textfeat and
\titlefeat. Participants performed their text analysis primarily on the \titlefeat or \textfeat fields, while additional features were available if needed. The task was to predict the labels \prodcat and \hazcat, as well as the vectors \prodvec and \hazvec. The dataset comprises 1,256 different \prodvec values (e.g.,
  ``\textit{ice cream},''
  ``\textit{chicken based products},''
  ``\textit{cakes}'') sorted into 22 categories (e.g.
  ``\textit{meat, egg and dairy products},''
  ``\textit{cereals and bakery products},''
  ``\textit{fruits and vegetables}'') with the help of ontologies.
In addition, there are 261 distinct values for \hazvec (e.g.,
  ``\textit{salmonella},''
  ``\textit{listeria monocytogenes},''
  ``\textit{milk and products thereof}'')
, which are grouped (again using ontologies) into the following 10 values of the label \hazcat:
  ``\textit{allergens},''                        
  ``\textit{biological},''                       
  ``\textit{foreign bodies},''                   
  ``\textit{fraud},''                            
  ``\textit{chemical},''                         
  ``\textit{other hazard},''                     
  ``\textit{packaging defect},''                 
  ``\textit{organoleptic aspects},''             
  ``\textit{food additives and flavourings},''   
  ``\textit{migration}.''                        
The class distribution in the data is heavily imbalanced with the above examples being ranked from the most to the least common in Figure~\ref{fig:data-stats}.

\subsection{Baselines}
In our challenge, we provided participants with three jupyter-notebooks for training and evaluating baseline models for both subtasks\footnote{\href{https://github.com/food-hazard-detection-semeval-2025/food-hazard-detection-semeval-2025.github.io/tree/main/code}{\texttt{https://food-hazard-detection-semeval-2025.\\github.io/code/}}}:

\noindent\textbf{(i)}~We provide a traditional pipeline consisting of a TF-IDF embedding in combination with a logistic regression classifier based on the \texttt{scikit-learn} Python module~\cite{scikit-learn}.

\noindent\textbf{(ii)}~A second baseline implementation finetunes an encoder-only transformer, specifically \texttt{bert-base-uncased}~\cite{BERT}, using the \texttt{transformers} Python module~\cite{wolf2020huggingfacestransformers} by \href{https://huggingface.co/docs/transformers}{huggingface.co}.

\noindent\textbf{(iii)}~Finally we provide a more sophisticated baseline based on the CICLe method~\citet{randl2024cicle}. It relies on prompting larger transformers such as GPT-4 without further fine-tuning \cite{GPT} in combination with conformal prediction~\cite{Vovk_conformal}. In our baseline we use the \texttt{crepes} Python module to implement conformal prediction \cite{crepes}.

Baseline performance of different classifiers on the whole dataset used in this challenge was also reported by \citet{randl2024cicle}. The results showed that the classification of hazards and products was a non-trivial task, and the classification of the ``vector''-label, which we aimed to address in this challenge, was particularly challenging.

\section{Evaluation} \label{sec:eval}

We computed the performance for ST1 and ST2 by calculating the macro~$\text{\sc f}_1$-score on the participants' predicted labels $\hat{{\bf{y}}}$ using the annotated labels ${\bf{y}}$ as ground truth. This measure is the unweighted mean of per-class-$\text{\sc f}_1$-scores over the $n$ classes. Both $\hat{{\bf{y}}}$ and ${\bf{y}}$ are vectors of $m$ samples:
\begin{equation}
    \text{\sc f}_1({\bf{y}}, \hat{{\bf{y}}}) = \frac{2}{n} \sum_{i=0}^{n}{\frac{\text{\sc rcl}_i({\bf{y}}, \hat{{\bf{y}}}) \cdot \text{\sc prc}_i({\bf{y}}, \hat{{\bf{y}}})}{{\text{\sc rcl}_i({\bf{y}}, \hat{{\bf{y}}})} + \text{\sc prc}_i({\bf{y}}, \hat{{\bf{y}}})}}
\end{equation}
where $\text{\sc rcl}_c$ is the recall and $\text{\sc prc}_c$ is the precision for a specific class $c$.
In order to combine the predictions for the \hazvec and \prodvec labels into one score, we took the average of the scores:
\begin{equation}
    \text{\sc s}(Y, \hat{Y}) = \frac{\text{\sc f}_1({\bf{y}}^{h}, \hat{{\bf{y}}}^{h}) + \text{\sc f}_1({\bf{y}}^{p|h}, \hat{{\bf{y}}}^{p|h})}{2}
\end{equation}
Here $Y = [{\bf{y}}^{h}, {\bf{y}}^{p}]$ is the $2 \times m$ matrix with the \hazvec label~${\bf{y}}_{h}$ and the \prodvec label~${\bf{y}}_{p}$ as column vectors. The vector ${\bf{y}}^{p|h}$ is defined as the entries of ${\bf{y}}^{p}$ where ${\bf{y}}^{h}$ is correctly predicted:
\begin{equation}
    {\bf{y}}^{p|h} = \{{\bf{y}}_{j}^{p}~|~\hat{\bf{y}}_{j}^{h}={{\bf{y}}}_{j}^{h}\},~j \in \{1,2,...,m\}
\end{equation}
The scalar ${\bf{y}}_{j}^{*}$ is the $j$-th element of ${\bf{y}}^{*}$. $\hat{Y}$ and $\hat{\bf{y}}^{p|h}$ are defined accordingly.
With this measure we based our rankings predominantly on the predictions for the \hazvec classes. Intuitively, this means that a submission with both ${\bf{y}}^{h}$ and ${\bf{y}}^{p}$ completely right would have scored $1.0$, a submission with ${\bf{y}}^{h}$ completely right and ${\bf{y}}^{p}$ completely wrong would have scored $0.5$, and any submission with ${\bf{y}}^{h}$ completely wrong would have scored $0.0$ independently of the value of ${\bf{y}}^{p}$.

\section{Participant Systems and Results}
In total, our task attracted approximately 260 participants and received 99 valid submissions during the evaluation phase. Among these, 27 system description papers were authored and submitted for peer-review. These 27 systems form the basis of our analysis and the official ranking, as they are accompanied by detailed system descriptions, enabling a thorough evaluation. The full, unofficial ranking -- including all submissions to codalab -- is available on the task’s website.\footnote{\url{https://food-hazard-detection-semeval-2025.github.io/}}

\subsection{Popular Methods}

\begin{figure*}
    \centering
    \includegraphics[trim={0 50 0 60}, clip, width=\linewidth]{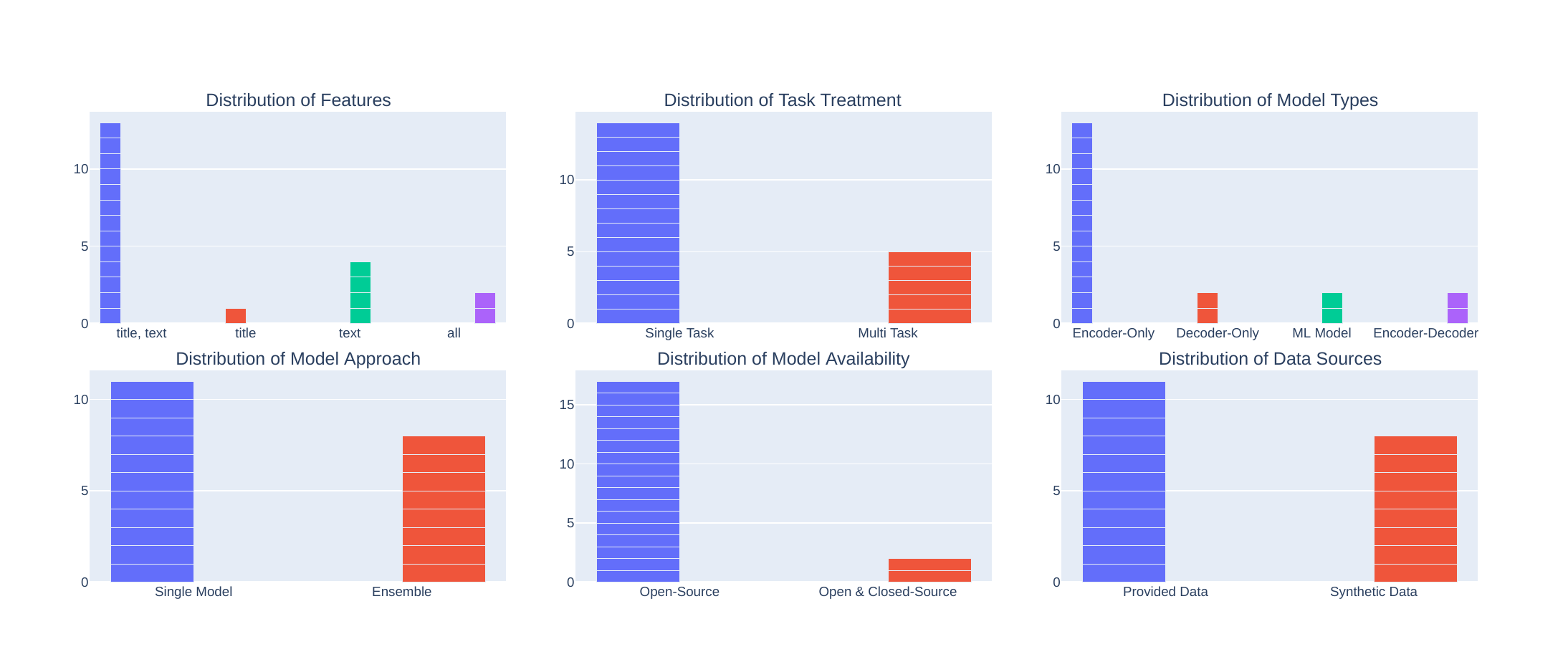}
    \caption{Frequency distribution of system attributes. Each subplot represents a distinct attribute, illustrating the choices made by the participating systems in terms of features, task treatment, model types, ensemble strategies, model availability, and data usage.}
    \label{fig:feat-distr}
\end{figure*}

Figure~\ref{fig:feat-distr} illustrates the frequency distribution of system attributes. Each subplot corresponds to a distinct attribute, highlighting key trends among the systems. We observe that all systems that addressed both subtasks employed the same features for each, with the majority (14 systems) using both \titlefeat and \textfeat
features, while two systems incorporated all available dataset features. Furthermore, the majority (14 systems) treated the tasks separately, with only five systems leveraging a combined approach to exploit the correlation between the tasks. In terms of model choice, most systems (13) relied on encoder-only Transformed-based models, while two used traditional machine learning models. Among the systems that used transformer-based models, open-source models (17 systems) were  preferred. Furthermore, the majority (11 models) opted for a single model for classification rather than an ensemble strategy (8 systems). Finally, regarding the data sources, eight systems incorporated synthetic data to address the tasks.

\subsection{Leaderboard Results}\label{sec:leaderboard}
\paragraph{ST1} Table~\ref{tab:leaderboard-st1} presents the results and the ranking of the systems that submitted system a description paper in ST1. The scores lie between 0.1426 and 0.8223, with the largest gap in performance observed between the first and second-ranked systems among the top three. Systems ranked between fifth and 16th exhibit relatively similar scores, while a distinct widening of the gap is evident in the lower ranks. Furthermore, the top two systems used richer feature sets compared to the lower-ranked systems, indicating that the richer feature sets may have contributed to their scores, while most systems relied on both textual features, i.e., \titlefeat and \textfeat, rather than focusing on one of them.

\begin{table}[h]
    \centering
    \resizebox{\linewidth}{!}{\begin{tabular}{rlrl}
    \toprule
    \textsc{Rank} & \textsc{Team Name} & \textsc{Score} & \textsc{Features} \\
    \midrule
    \multirow{2}{*}{\footnotesize Baselines:}
    & {\footnotesize TFIDF + LR} & {\footnotesize 0.498} & {\footnotesize \titlefeat} \\
    & {\footnotesize BERT      } & {\footnotesize 0.667} & {\footnotesize \titlefeat} \\
    \midrule
    \multirow{2}{*}{ 1} & \multirow{2}{*}{Anastasia}  & \multirow{2}{*}{0.8223} & \footnotesize \yearfeat, \monthfeat, \dayfeat,\\
                        &                             &                         & \footnotesize \country, \titlefeat, \textfeat \\
    \multirow{2}{*}{ 2} & \multirow{2}{*}{MyMy}       & \multirow{2}{*}{0.8112} & \footnotesize \yearfeat, \monthfeat, \dayfeat,\\
                        &                             &                         & \footnotesize \country, \titlefeat, \textfeat \\
                   { 3} & SRCB                        &                {0.8039} & \footnotesize \titlefeat, \textfeat \\
                   { 4} & PATeam                      &                {0.8017} & \footnotesize \titlefeat, \textfeat \\
                   { 5} & HU                          &                {0.7882} & \footnotesize \titlefeat, \textfeat \\
                   { 6} & BitsAndBites                &                {0.7873} & \footnotesize \titlefeat, \textfeat \\
                   { 7} & CSECU-Learners              &                {0.7863} & \footnotesize \titlefeat, \textfeat \\
                   { 8} & ABCD                        &                {0.7860} & \footnotesize \titlefeat, \textfeat \\
                   { 9} & MINDS                       &                {0.7857} & \footnotesize \titlefeat, \textfeat \\
                   {10} & Zuifeng                     &                {0.7835} & \footnotesize N/A \\
                   {11} & Fossils                     &                {0.7815} & \footnotesize \titlefeat, \textfeat \\
                   {12} & PuerAI                      &                {0.7729} & \footnotesize N/A \\
                   {13} & Ustnlp16                    &                {0.7654} & \footnotesize \titlefeat, \textfeat \\
                   {14} & FuocChu\_VIP123             &                {0.7646} & \footnotesize N/A \\
                   {15} & BrightCookies               &                {0.7610} & \footnotesize \textfeat \\
                   {16} & farrel\_dr                  &                {0.7587} & \footnotesize \titlefeat, \textfeat \\
                   {17} & OPI-DRO-HEL                 &                {0.7381} & \footnotesize \titlefeat, \textfeat \\
                   {18} & madhans476                  &                {0.7362} & \footnotesize \titlefeat, \textfeat \\
                   {19} & Anaselka                    &                {0.6858} & \footnotesize \titlefeat, \textfeat \\

    {\color{gray}\multirow{2}{*}{20}} & {\color{gray}\multirow{2}{*}{Somi}      } & {\color{gray}\multirow{2}{*}{0.6614}} & {\color{gray}\footnotesize \titlefeat, \textfeat} \\
                                      &                                           &                                       & {\color{gray}\footnotesize \country, \titlefeat, \textfeat} \\
    {\color{gray}               {21}} & {\color{gray}TechSSN3                   } & {\color{gray}               {0.6442}} & {\color{gray}\footnotesize \textfeat} \\
    {\color{gray}               {22}} & {\color{gray}UniBuc                     } & {\color{gray}               {0.6355}} & {\color{gray}\footnotesize \titlefeat, \textfeat} \\
    {\color{gray}               {23}} & {\color{gray}CICL                       } & {\color{gray}               {0.6079}} & {\color{gray}\footnotesize \textfeat} \\
    {\color{gray}               {24}} & {\color{gray}VerbaNexAI                 } & {\color{gray}               {0.5165}} & {\color{gray}\footnotesize \titlefeat} \\
    {\color{gray}               {25}} & {\color{gray}JU-NLP                     } & {\color{gray}               {0.4566}} & {\color{gray}\footnotesize \titlefeat, \textfeat} \\
    {\color{gray}               {26}} & {\color{gray}Habib University           } & {\color{gray}               {0.4482}} & {\color{gray}\footnotesize N/A} \\
    {\color{gray}               {27}} & {\color{gray}Howard University-AI4PC    } & {\color{gray}               {0.1426}} & {\color{gray}\footnotesize \textfeat} \\
    \bottomrule
    \end{tabular}}
    \caption{ST1 ranking for systems of teams that submitted a system description paper. Gray entries are outperformed by the best baseline.}
    \label{tab:leaderboard-st1}
\end{table}

\noindent\paragraph{ST2} Table~\ref{tab:leaderboard-st2} presents the results and the rankings of the systems for ST2 that submitted a system description paper. The results show significantly lower performance compared to ST1, with the highest score of SRCB (0.5473) being considerably lower than the top score in ST1 (0.8223), which indicates that ST2 is a more challenging task. A sharp drop in scores is observed after the top three teams and again after the 12th team (BitsAndBites), with the lowest-ranked system (Anaselka) receiving 0.0049 score. Notably, the top three teams in both subtasks -- except for the Anastasia team, which focused only on ST1 -- performed well in both, consistently ranking among the top teams in each subtask. Among the top 15 systems, while a few systems, such as Anastasia and BitsAndBites, performed better on ST1, a larger number of systems, including MINDS, Fossils, PuerAI, and BrightCookies, achieved significantly higher rankings in ST2.

\begin{table}[H]
    \centering
    \resizebox{\linewidth}{!}{\begin{tabular}{rlrl}
    \toprule
    \textsc{Rank} & \textsc{Team Name} & \textsc{Score} & \textsc{Features} \\
    \midrule
    \multirow{2}{*}{\footnotesize Baselines:}
    & {\footnotesize TFIDF + LR} & {\footnotesize 0.183} & {\footnotesize \titlefeat} \\
    & {\footnotesize BERT      } & {\footnotesize 0.165} & {\footnotesize \titlefeat} \\
    \midrule
                   { 1} & SRCB                        &                {0.5473} & \footnotesize \titlefeat, \textfeat \\
    \multirow{2}{*}{ 2} & \multirow{2}{*}{MyMy}       & \multirow{2}{*}{0.5278} & \footnotesize \yearfeat, \monthfeat, \dayfeat,\\
                        &                             &                         & \footnotesize \country, \titlefeat, \textfeat \\
                   { 3} & PATeam                      &                {0.5266} & \footnotesize \titlefeat, \textfeat \\
                   { 4} & HU                          &                {0.5099} & \footnotesize \titlefeat, \textfeat \\
                   { 5} & MINDS                       &                {0.4862} & \footnotesize \titlefeat, \textfeat \\
                   { 6} & Fossils                     &                {0.4848} & \footnotesize \titlefeat, \textfeat \\
                   { 7} & CSECU-Learners              &                {0.4797} & \footnotesize \titlefeat, \textfeat \\
                   { 8} & PuerAI                      &                {0.4783} & \footnotesize N/A \\
                   { 9} & Zuifeng                     &                {0.4712} & \footnotesize N/A \\
                   {10} & ABCD                        &                {0.4576} & \footnotesize \titlefeat, \textfeat \\
                   {11} & BrightCookies               &                {0.4529} & \footnotesize \textfeat \\
                   {12} & Ustnlp16                    &                {0.4512} & \footnotesize \titlefeat, \textfeat \\
                   {13} & BitsAndBites                &                {0.4456} & \footnotesize \titlefeat, \textfeat \\
                   {14} & UniBuc                      &                {0.3453} & \footnotesize \titlefeat, \textfeat \\
                   {15} & OPI-DRO-HEL                 &                {0.3295} & \footnotesize \titlefeat, \textfeat \\
                   {16} & VerbaNexAI                  &                {0.3223} & \footnotesize \titlefeat \\
                   {17} & CICL                        &                {0.3169} & \footnotesize \textfeat \\
    \multirow{2}{*}{18} & \multirow{2}{*}{Somi}       & \multirow{2}{*}{0.3048} & \footnotesize \titlefeat, \textfeat \\
                        &                             &                         & \footnotesize \country, \titlefeat, \textfeat \\
                   {19} & TechSSN3                    &                {0.2712} & \footnotesize \textfeat \\

    {\color{gray}               {20}} & {\color{gray}Howard University-AI4PC    } & {\color{gray}               {0.1380}} & {\color{gray}\footnotesize \textfeat} \\
    {\color{gray}\multirow{2}{*}{21}} & {\color{gray}\multirow{2}{*}{Anastasia} } & {\color{gray}\multirow{2}{*}{0.1281}} & {\color{gray}\footnotesize \yearfeat, \monthfeat, \dayfeat,}\\
                                      &                                           &                                       & {\color{gray}\footnotesize \country, \titlefeat, \textfeat} \\
    {\color{gray}               {22}} & {\color{gray}farrel\_dr                 } & {\color{gray}               {0.1249}} & {\color{gray}\footnotesize \titlefeat, \textfeat} \\
    {\color{gray}               {23}} & {\color{gray}madhans476                 } & {\color{gray}               {0.0486}} & {\color{gray}\footnotesize \titlefeat, \textfeat} \\
    {\color{gray}               {24}} & {\color{gray}Habib University           } & {\color{gray}               {0.0315}} & {\color{gray}\footnotesize N/A} \\
    {\color{gray}               {25}} & {\color{gray}JU-NLP                     } & {\color{gray}               {0.0126}} & {\color{gray}\footnotesize \titlefeat, \textfeat} \\
    {\color{gray}               {26}} & {\color{gray}Anaselka                   } & {\color{gray}               {0.0049}} & {\color{gray}\footnotesize \titlefeat, \textfeat} \\
    \bottomrule
    \end{tabular}}
    \caption{ST2 ranking for systems of teams that submitted a system description paper. Gray entries are outperformed by the best baseline.}
    \label{tab:leaderboard-st2}
\end{table}

\subsection{Best Systems}\label{sec:best}

In this section, we outline the key methods employed by the top three systems for each subtask. Since the teams ``MyMy'' and ``SCRB'' rank among the top three in both evaluations (see Tables~\ref{tab:leaderboard-st1} and \ref{tab:leaderboard-st2}), we analyze a total of four systems.

\paragraph{SRCB~{\footnotesize(ST1:~$3^{rd}$, ST2:~$1^{st}$)}}
The first place in ST2 comes from \href{https://www.ricoh.com/}{Ricoh Software Research Center}. \citet{submission225} concatenated the \titlefeat and the \textfeat in lower case, and followed a two-step approach. In the first step, they used BERT to reduce the label space and include only the most probable ones. In a second step, then, all the possible labels (possibly along with examples) were fed to a large language model~(LLM) to predict the correct one. This approach follows the paradigm of \citet{randl2024cicle}, who suggested reducing the possible labels by quantifying the uncertainty with conformal prediction~\cite{Vovk_conformal}. Infrequent categories were furthermore augmented with an LLM, while approx. 10\% of the data was truncated.  

\paragraph{Anastasia~{\footnotesize(ST1:~$1^{st}$, ST2:~$21^{th}$)}} The best system in ST1 is by \citet{submission102} from \href{https://en.uit.edu.vn/}{VNUHCM -- University of Information Technology} and focuses on ST1. After a simple text normalization step, they chunk the texts into snippets of consecutive sentences that fit the context windows of their applied models. Following this, they fine-tune two encoder-only transformers, specifically \texttt{DeBERTa-v3-large} and \texttt{RoBERTa-large}, using focal loss~\citep{Lin2017_FocalLoss} with class weights. For training, they compare two setups:
\textbf{(i)}~They try \textbf{multi-task} fine-tuning of \texttt{DeBERTa-v3-large} to get a combined model for both \hazvec and \prodvec prediction using oversampling for underrepresented classes and undersampling for overrepresented classes.
\textbf{(ii)}~Additionally, they try \textbf{single-task} fine-tuning of both \texttt{DeBERTa-v3-large} and \texttt{RoBERTa-large}, this time addressing the class-imbalance by creating synthetic samples by prompting \texttt{gemini-2.0-flash-exp} to paraphrase texts in underrepresented classes.
They report that multi-task training leads to slightly worse performance on ST1 compared to single-task. This may be owed to the different resampling approaches, though.
Finally, they combine all of their trained models (single- and multi-task) in one ensemble, using soft voting with a weighted sum. The weights were based on grid search on the validation set.

\begin{figure*}[t!]
    \centering
    \includegraphics[width=.8\linewidth]{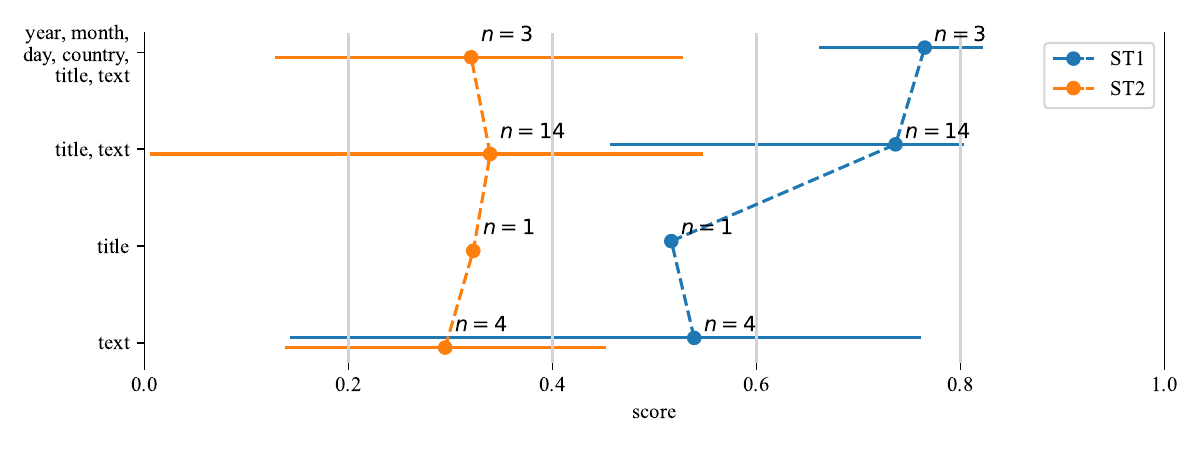}
    \caption{Average score achieved and number of submissions per combination of input features used ({\color{blue}--}~ST1, {\color{orange}--}~ST2). The horizontal bars show minimum and maximum score and the number of samples is annotated as $n$.}
    \label{fig:features}
\end{figure*}

\paragraph{MyMy~{\footnotesize(ST1:~$2^{nd}$, ST2:~$2^{nd}$)}} \citet{submission204} from the \href{https://www.csie.ncku.edu.tw/en/}{Department of Computer Science and Information Engineering, National Cheng Kung University} employ a retrieval-augmented generation (RAG) approach to address both subtasks separately by intergrating domain-specific external knowledge. It first retrieves relevant documents for each data sample from PubMed,\footnote{\url{https://pubmed.ncbi.nlm.nih.gov/}} following the RAG paradigm: it uses GPT- 3.5 Turbo3,\footnote{\url{https://platform.openai.com/docs/models/ gpt-3-5-turbo}} Gemini Flash 2.0 \cite{team2023gemini}, Llama 3.1 8B \cite{touvron2023llama}, and Mistral 8x7B \cite{jiang2023mistral7b} LLMs to simplify the original data sample; it then retrieves documents from the PubMed API, encodes them into embeddings using \texttt{nomic-embed-text-v1} \cite{nussbaum2024nomic} and stores them in a Chroma embedding database; cosine similarity scores are then computed to retrieve the top-K most relevant documents. These documents are then combined with the original input and paraphrased using the same LLMs to generate augmented data. A validation step incorporating the same LLMs is used to filter the generated samples based on relevance, ensuring data quality. The enriched dataset is then used to fine-tune classification models (Gemini Flash 2.0 \cite{team2023gemini}, PubMedBERT \cite{gu2021domain}, and ModernBERT \cite{warner2024smarter}). Finally, predictions are obtained through a weighted soft voting strategy, where class probabilities from multiple models are combined using weighted sums to determine the final label.

\paragraph{PATeam~{\footnotesize(ST1:~$4^{th}$, ST2:~$3^{rd}$)}} \citet{submission354} begin with data cleaning using regular expressions, followed by text augmentation, where LLM-generated summaries are concatenated with the \textfeat feature. To address data imbalance, SMOTE \cite{chawla2002smote} is applied to underrepresented categories (fewer than five samples, a threshold determined through tuning) to ensure a minimum of five samples per class. The system employs a bagging approach with bootstrapping to generate five subsets of the training data, fine-tuning five \texttt{microsoft/phi-4 models}\footnote{\url{https://huggingface.co/microsoft/phi-4}} using low-rank adaptation (LoRA) \cite{hu2021loralowrankadaptationlarge} to reduce trainable parameters. Predictions from all five models are integrated via an ensemble voting mechanism. The system employs the multi-dimensional type-slot label interaction network (MTLN) \cite{wan2023unified} to capture the correlation between the two subtasks. It first classifies ST1 and then utilizes these predictions to inform the classification of ST2. An ablation study confirmed that this multi-task approach outperforms treating the tasks independently.

\subsection{What Worked Well}

A prevalent strategy among these systems is the use of generative LLMs for synthetic data creation to mitigate class imbalance. Specifically, three approaches stand out: (i)~\textbf{paraphrasing}~\citep{submission102}, (ii)~\textbf{summarizing} and appending generated text to the original~\citep{submission354}, and (iii)~\textbf{generating} new samples by combining information from two instances of the same class~\citep{submission225}. Additionally, \citet{submission102} and \citet{submission204} incorporate class-weighted loss functions to increase the impact of underrepresented classes during training.

Another common technique among top-ranking systems is the use of ensemble methods. \citet{submission102} employ a soft voting approach, optimizing weights of different models during grid search on a validation set, while \citet{submission204} adopt a max voting strategy, selecting the prediction from the most confident model. \citet{submission354} fine-tune five classifiers using bootstrapped subsets of their preprocessed training data.

In contrast to these shared strategies, there is no clear consensus on the use of multi-task learning (joint modeling of both subtasks) versus single-task learning (treating subtasks separately). Three out of four systems opt for a single-task approach, but \citet{submission354} experiment with both strategies in a prompting-based classification setup. Their results suggest that multi-turn prompts, where both subtasks are addressed within a single interaction, outperform single-turn prompts, which handle the subtasks separately.

As discussed in Section~\ref{sec:leaderboard}, richer feature sets tend to support stronger models across both subtasks. This observation is further illustrated in Figure~\ref{fig:features}: while using either \textfeat or \titlefeat alone does not appear to significantly improve performance compared to the other, combining both leads to improvements in both subtasks. However, the limited number of approaches that rely solely on \titlefeat makes it difficult to draw definitive conclusions. A similar challenge arises for the two approaches that utilize all available features: although they achieve better results in ST1, they underperform in ST2, suggesting that their design was primarily optimized for the former.

\begin{figure}[H]
    \centering
    \includegraphics[width=\linewidth]{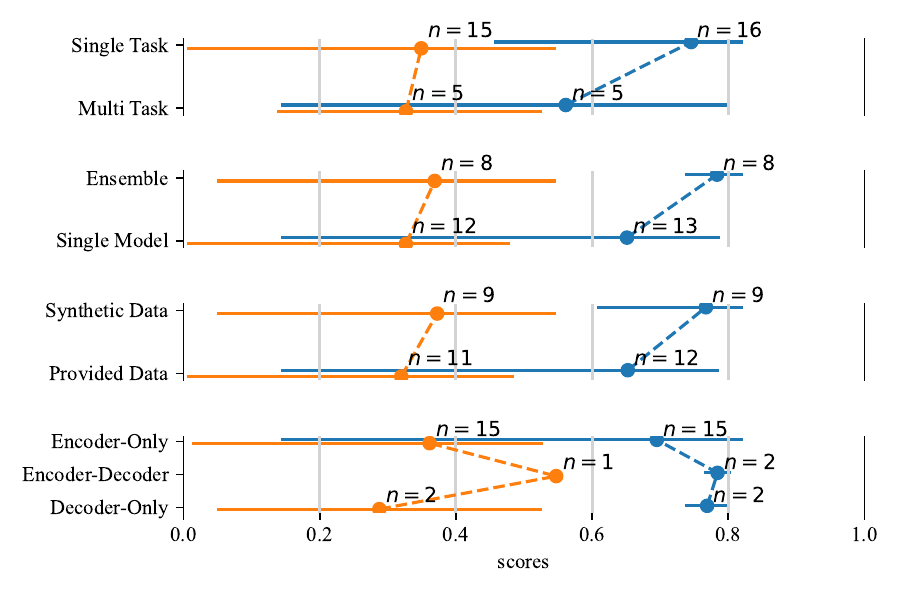}
    \caption{Average score achieved and number of submissions per combination per design choice ({\color{blue}--}~ST1, {\color{orange}--}~ST2). The horizontal bars show minimum and maximum.}
    \label{fig:comparison}
\end{figure}

Figure~\ref{fig:comparison} presents a detailed comparison of design choices based on evaluation scores. Interestingly, treating the subtasks separately leads to better performance than multi-task approaches that use a shared model for ST1 and ST2. Additionally, leveraging an ensemble of multiple models proves more effective than relying on independent models.

As discussed in Section~\ref{sec:data}, one major challenge participants faced was the extreme class imbalance in the dataset. It is therefore unsurprising that oversampling underrepresented classes with artificially generated data significantly improved performance compared to using only the provided training set. As noted in Section~\ref{sec:best}, this artificial data was typically generated by prompting LLMs.
Finally, an interesting finding is that no transformer architecture -- whether encoder-only (e.g. BERT), encoder-decoder (e.g. BART), or decoder-only (e.g. Llama) -- consistently outperforms the others. Across all three architectures, the highest achieved scores remain approximately equal within each subtask.

\section{Discussion}
\subsection{Task difficulty estimation}
\begin{figure}
    \centering
    \includegraphics[width=\linewidth]{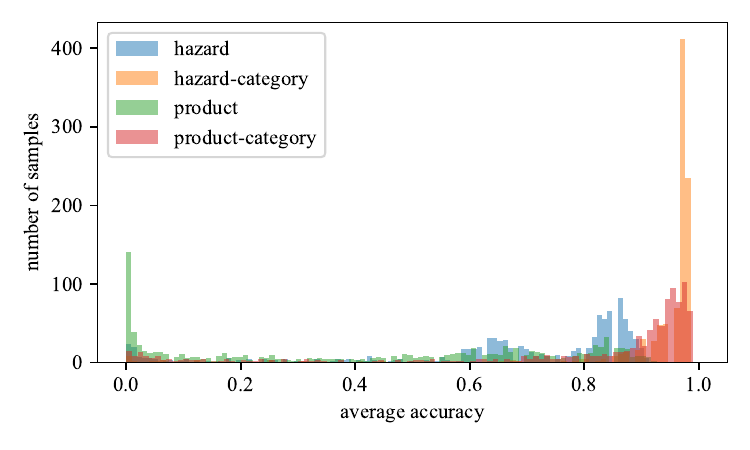}
    \caption{Histogram of the frequency (vertically) across the fraction of systems correctly predicting a specific sample (horizontally).}
    \label{fig:sample_difficulty}
\end{figure}

We show an instance-based difficulty analysis in Figure~\ref{fig:sample_difficulty}. The figure shows that across categories/vectors most samples are more likely to be predicted correctly than not. Nevertheless, we also see a spike at zero accuracy, which is most prevalent for the vector \prodvec, but seen for all categories/vectors. This indicates that several samples were never correctly classified, indicating that they are extremely difficult or even missing information. To make it easier to identify such instances in our data, we include an instance difficulty score, ranging linearly from 0~(\textit{instance was classified correctly by all submissions}) to 1~(\textit{instance was never correctly classified}), for all instances in the train and test set in our dataset on zenodo.

\subsection{Error Analysis}
\begin{figure}[H]
    \centering
    \includegraphics[width=\linewidth]{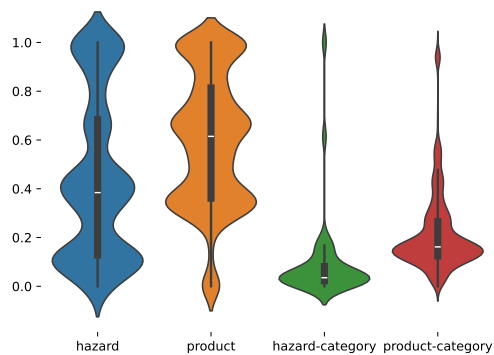}
    \caption{Pairwise error rate of submissions}
    \label{fig:error_analysis}
\end{figure}

Figure~\ref{fig:error_analysis} shows the pairwise error rate between the submissions per category. The error is considerably higher in ST2 for the two plots on the right compared to the two of ST1. This is partly due to the fewer number of possible labels in the latter and the higher likelihood of mistakes on the former.

A more detailed analysis is shown in the confusion matrices in Appendix~\ref{app:confusion}. For \hazcat, we see that precision and recall are relatively high except for the classes ``\textit{migration}'' and ``\textit{food additives and flavourings}'' (see Figure~\ref{fig:confusion_hazard-category}). While samples of the class ``\textit{migration}'' are predicted to the very similar class ``\textit{chemical}'' in 90\% of the cases, predictions for ``\textit{food additives and flavourings}'' are divided between the true class (49\%), ``\textit{other hazard}'' (28\%), ``\textit{fraud}'' (13\%), and ``\textit{allergens}'' (13\%).

We see a similar picture for \prodcat in Figure~\ref{fig:confusion_product-category}. Most classes show good performance, while ``\textit{food additives and flavourings},'' ``\textit{honey and royal jelly},'' and ``\textit{other food product / mixed}'' show high misclassification rates. ``\textit{Food additives and flavourings}'' is most commonly confused with ``\textit{meat, egg and dairy products}'' (22\%) and ``\textit{cereals and bakery products}'' (18\%). ``\textit{Honey and royal jelly}'' is confused with the most supported class ``\textit{meat, egg and dairy products}'' in 40\% of the cases. As an overarching class for leftover samples, ``\textit{other food product / mixed}'' is misclassified to multiple other classes, most prominently ``\textit{soups, broths, sauces and condiments}'' (18\%), and ``\textit{fruits and vegetables}'' (18\%).
All of these commonly mislabeled classes are highly underrepresented in the dataset and/or easy to confuse with other, higher-supported classes in the data.

\section{Conclusion}
In conclusion, our task demonstrates that LLM-generated synthetic data can be highly effective for oversampling in long-tail distributions. A second, albeit expected, finding is that ensemble strategies significantly enhance classification performance. Additionally, while combined approaches for vector and category classification can be beneficial in prompting scenarios, they do not generally lead to performance improvements. More notably, we do not observe a clear winner among transformer architectures: fine-tuned encoder-only, encoder-decoder, and decoder-only models achieve comparable maximum performance across both subtasks.

Future research on our dataset should prioritize the more challenging vector classification task. Our analysis indicates that classification errors often stem from low class support and that food recall texts contain ambiguous instances, with semantically similar classes contributing to misclassification. We argue that debugging classifiers using explainability techniques may help improve performance.

Despite its potential to assist human validation and enable meta-learning approaches, such as clustering or pre-sorting examples, explainability in text-based food risk classification remains underexplored. However, explanations can vary significantly depending on the model and task. Existing literature addresses both model-specific~\cite{19, 20} and model-agnostic~\cite{21} explainability approaches, which should be further investigated in this domain.

\section*{Limitations}
\noindent\textbf{(i)}~A limitation of our evaluation process is that, while we enforced a one-submission-per-user policy during the evaluation phase, some participants have circumvented this by registering multiple accounts. We chose not to remove suspicious accounts, as identifying all of them would have been impractical and likely only encouraged more covert attempts to bypass the restriction.

\noindent\textbf{(ii)}~We chose to release the unlabeled test set at the beginning of the challenge, as this was easier to set up with codalab. While this ensured transparency throughout the challenge, participants strongly determined to win could peak (e.g., manually annotating the test data).

\noindent\textbf{(iii)}~We found 42 duplicate entries in our dataset after the start of the challenge. These were introduced due to an error in one of our preprocessing scripts and resulted in 6 entries that are present in both the training and validation set as well as 7 entries that are present in both the training and test set. As this concerns less than 1\% of the data, we argue that it is not severely impacting our results.

\section*{Ethical statement}
All texts are collected from official and publicly available sources, hence no privacy-related issues are present. All annotations have been provided by \href{https://agroknow.com/}{Agroknow} experts. System application is intended to complement and not substitute the human expert in preventing illness or harm from food sources.

\section*{Acknowledgments}

We thank \href{mailto:stoitsis@agroknow.com}{Giannis Stoitsis} and \href{https://agroknow.com}{Agroknow} for collecting and providing the data for this task.

This work has been partially funded by the European Union under the NextGenerationEU program as part of project MIS 5154714 of the National Recovery and Resilience Plan Greece 2.0. 

This research has also been partially funded by the European Union’s Horizon Europe research and innovation program EFRA (Grant Agreement Number 101093026).
Views and opinions expressed are however those of the authors only and do not necessarily reflect those of the European Union or European Commission-EU. Neither the European Union nor the granting authority can be held responsible for them. {\normalsize\euflag}

\bibliography{foodrisk}

\newpage
\appendix

\section{Confusion Matrices}
\label{app:confusion}

\begin{figure}[H]
    \centering
    \includegraphics[width=\linewidth]{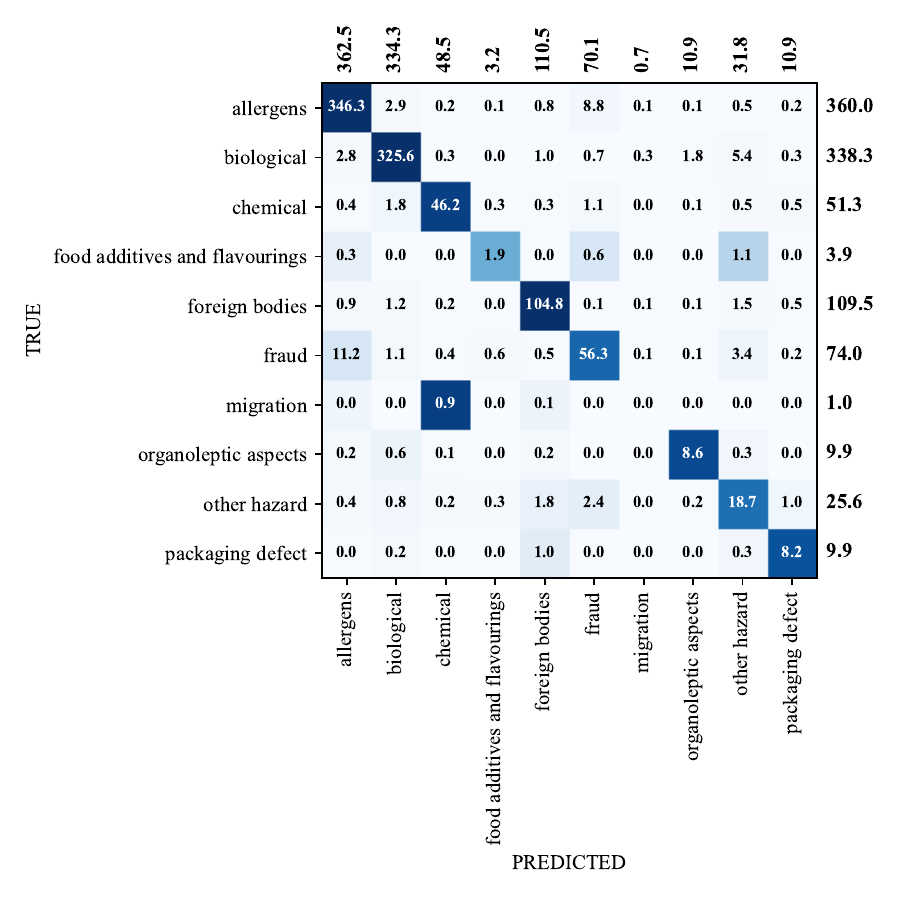}
    \caption{Confusion Matrix for \hazcat. Numbers signify average number of occurrences per submission during the evaluation phase. Colors are normalized by row.}
    \label{fig:confusion_hazard-category}
\end{figure}

\begin{figure*}[p]
    \centering
    \includegraphics[width=\linewidth]{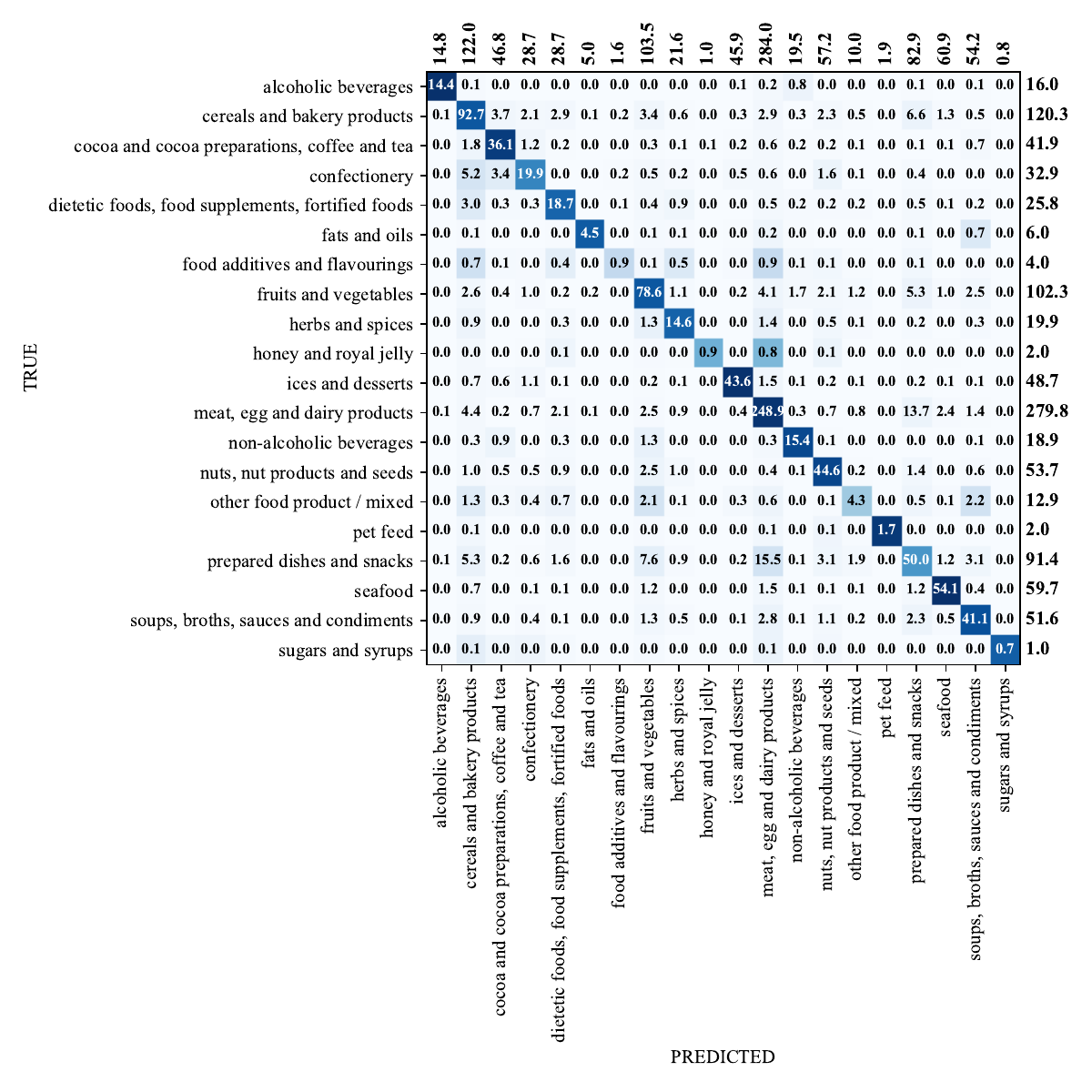}
    \caption{Confusion Matrix for \prodcat. Numbers signify average number of occurrence per submission during the evaluation phase. Colors are normalized by row.}
    \label{fig:confusion_product-category}
\end{figure*}

\end{document}